\documentclass[a4paper, 10pt, conference]{IEEEtran}
\IEEEoverridecommandlockouts
\usepackage{enumitem}
\usepackage{hyperref}
\hypersetup{
    colorlinks=true,
    linkcolor=blue,
    filecolor=blue,
    citecolor=blue,
    urlcolor=blue,
    }

\usepackage{subfigure}
\usepackage{multirow}
\usepackage{makecell}
\usepackage{longtable}
\usepackage{color}
\usepackage[table]{xcolor}
\usepackage{authblk}

\usepackage{cite}
\usepackage{amsmath,amssymb,amsfonts}
\usepackage{algorithmic}
\usepackage{graphicx}
\usepackage{textcomp}
\usepackage{xcolor}
\def\BibTeX{{\rm B\kern-.05em{\sc i\kern-.025em b}\kern-.08em
    T\kern-.1667em\lower.7ex\hbox{E}\kern-.125emX}}
\begin{document}

\title{A Skeleton-aware Graph Convolutional Network for Human-Object Interaction Detection 
\thanks{Emails: manli.zhu@northumbria.ac.uk, e.ho@northumbria.ac.uk, hubert.shum@durham.ac.uk

* Corresponding author
}

}

\author[1]{Manli Zhu}
\author[1]{Edmond S. L. Ho}
\author[2*]{Hubert P. H. Shum}
\affil[1]{\small Department of Computer and Information Sciences, Northumbria University, Newcastle upon Tyne, UK} 
\affil[2]{\small Department of Computer Science, Durham University, Durham, UK}

\maketitle

\begin{abstract}
Detecting human-object interactions is essential for comprehensive understanding of visual scenes. In particular, spatial connections between humans and objects are important cues for reasoning interactions. To this end, we propose a skeleton-aware graph convolutional network for human-object interaction detection, named SGCN4HOI. Our network exploits the spatial connections between human keypoints and object keypoints to capture their fine-grained structural interactions via graph convolutions. It fuses such geometric features with visual features and spatial configuration features obtained from human-object pairs. Furthermore, to better preserve the object structural information and facilitate human-object interaction detection, we propose a novel skeleton-based object keypoints representation. The performance of SGCN4HOI is evaluated in the public benchmark V-COCO dataset. Experimental results show that the proposed approach outperforms the state-of-the-art pose-based models and achieves competitive performance against other models.
\end{abstract}

\begin{IEEEkeywords}
human-object interaction, graph convolutional network, deep learning, human pose, object skeleton
\end{IEEEkeywords}

\section{Introduction}
%background
Human-object interaction (HOI) detection is a core problem for in-depth understanding of visual scenes. The task requires localizing instances (humans and objects) and predicting their interactions in the form of $\left<human, action, object\right>$ from a given image. It plays an important role in analyzing visual scenes, such as visual question answering \cite{VQA} and activity analysis \cite{ActivityNet}. 

% transformer-based and non-transformer based, why do non-transformer 
Recent HOI detection approaches can roughly be divided into transformer-based methods \cite{HOITransformer,HOTR,QPIC} and non-transformer methods\cite{VSRL,IPNet,SG2HOI}. Particularly, the transformer-based models have shown superior performance, yet they require a large amount of memory and are difficult to train under limited resources. %As they often optimize object detection and interaction detection simultaneously, . 
As there has been significant progress in object detection such as Faster R-CNN \cite{Faster-R-CNN} and DETR \cite{detr}, many HOI models \cite{VSGNet,iCAN} utilized them to simplify the HOI detection task, resulting in a good trade-off between performance and complexity. However, as a human can interact with multiple objects and vice-versa, it remains challenging for these models to detect HOIs from visual image features. %First, a human can interact with multiple objects and vice-versa, making it difficult to identify their various roles in HOI triplets. Second, 
Human-object pairs with different interactions may share indistinguishable appearance or spatial configuration, which confuses classifiers that use such features.

% existing skeleton works and limitation
The advantage of modelling the mutual contexts of human pose and object in addition to their co-occurrence has been demonstrated in the literature \cite{Yao:CVPR2010}.  
%Human pose which provides fine-grained spatial cues for HOI detection has been studied. 
Most existing pose-based methods \cite{TIN,PMFNet,ETAI2022} simply use convolutional networks for human poses embedding, this can be inadequate for capturing the fine-grained human skeleton structure that is useful for HOI detection. For example, the ``foot" and ``legs" should be paid more attention than other human body parts in the scene of ``kick football". To address this, Zheng et al. \cite{SIGN} propose to use graph convolutional networks (GCN) for modelling the structured connections in human skeletons and the fine-grained interactions between human keypoints and object keypoints. They simply use two corner points of the object bounding box, however, either the 2-d points or 4-d bounding box representation considers only the rectangular spatial scope of an object and does not account for the shape and pose of semantically important local areas \cite{RepPoints,bbx-seg}.  %leading to an insufficient representation of real-world objects which have various sizes and shapes.

% our first contribution, human skeleton + object skeleton
In this paper, we propose a skeleton-aware graph convolutional network that makes use of both human skeleton keypoints and object skeleton keypoints for HOI detection, namely SGCN4HOI. Our idea is to utilize the keypoints from skeletons of humans/objects to capture their structural connections and as a guide to differentiating HOIs with subtle appearances and different structures. To this end, the graph convolutional network is exploited by considering keypoints of instances as nodes to capture their fine-grained spatial interactions, and we refer to this network as a {\it spatial skeleton stream}. In addition, to make full use of the available features, we add a visual stream that learns visual features from images and a spatial configuration stream that learns interaction patterns of human-object pairs to the framework. The two streams have proven to be effective and are commonly used in existing works \cite{VSRL,iCAN,VSGNet}. The three streams form our final multi-stream framework for HOI detection.

% difficulty of object representation and our solution
We also propose a novel skeleton-based method for representing object keypoints due to the limited literature on object skeleton and keypoints representation in HOI detection. Unlike humans that often have fixed skeleton structure, different kinds of objects usually have different structures, thus it is difficult to apply a unified algorithm for representing them. The dominant method for representing objects in object detection and HOI detection is object bounding box, which is too coarse to represent object's structure. Recent research such as \cite{CFA} propose to replace the bounding box with a convex hull consisting of 9 points for object representation, yet the main focus of their work is to make the object area more accurate and it is still supervised by the bounding box. Han et al. \cite{ReDet} propose to encode the coarse-grained orientation information for aerial object representation, while the object structure is not explored. Here, we propose to represent fine-grained object structure information in an indirect manner, i.e., we exploit a morphological skeletonization approach \cite{Zhang:skeleton84}  %(provided in Python library) 
that works on binary masks to obtain objects' skeletons. We extract a set of keypoints whose distribution can represent the objects' structural information using the K-means algorithm. %We extract key-point sets of these skeletons and apply the K-means algorithm to obtain a certain number of keypoints whose distribution can represent the objects' structural information.

\begin{figure*}[htbp]
	\setlength{\abovecaptionskip}{5pt} 
	\centering
	\includegraphics[width=\textwidth]{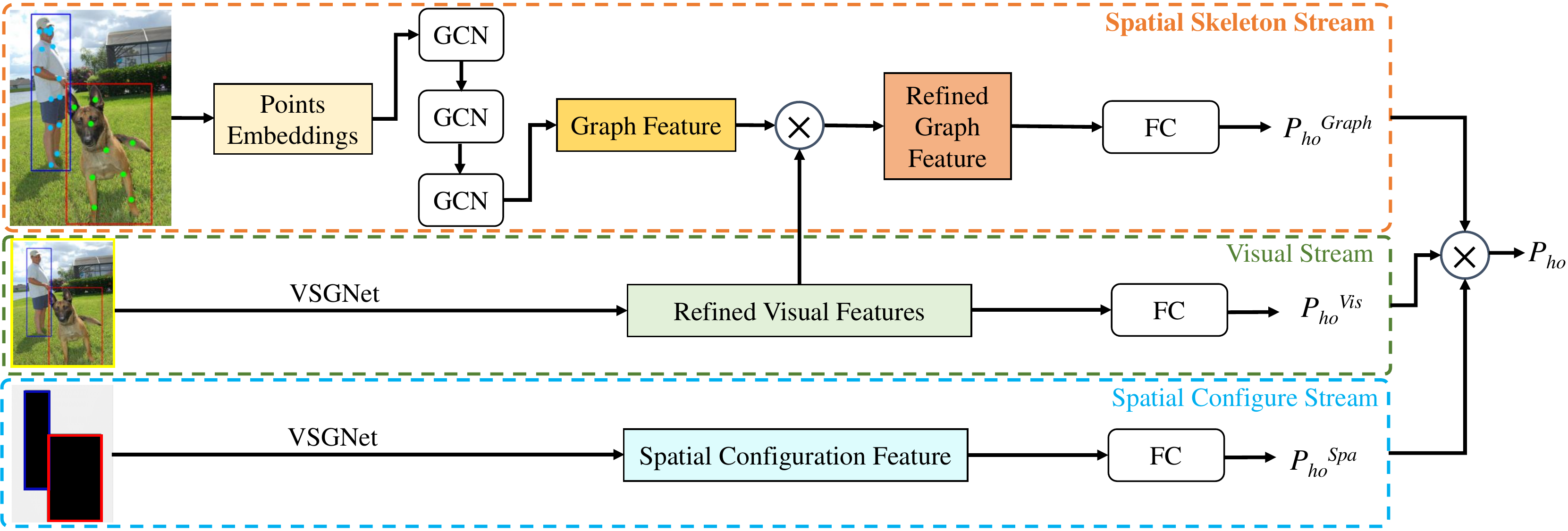}
	\caption{\footnotesize{Overview of the framework. Rounded rectangles represent operations, sharp rectangles with colors represent features, and $\otimes$ denotes element-wise multiplication. The model consists of three main streams. The Spatial configuration stream that extracts coarse-grained spatial features between humans and objects by utilizing the configuration of their bounding boxes and visual stream that extracts visual features from given images. The spatial skeleton stream extracts fine-grained spatial interaction features by modelling human keypoints and object keypoints as nodes and similarities between nodes plus learnable weights as edges. HOI probabilities from each stream are multiplied together as the final prediction.}}
	\label{Fig: framework}
\end{figure*}

%Evaluation
We evaluate our model in the HOI detection benchmark V-COCO dataset. Our proposed SGCN4HOI achieves state-of-the-art performance on this dataset compared to pose-based HOI models and other non-transformer methods. 

The contributions of this work are summarized as follows:
\begin{itemize}
    \item We propose a skeleton-aware graph network that considers the skeletons of both humans and objects to model their fine-grained spatial interactions.
    \item We propose a novel skeleton-based object keypoints representation method to preserve object structural information. To the best of our knowledge, this is the first work that exploits object structure representation in HOI detection.
    \item We demonstrate the effectiveness of our proposed model by conducting experiments in the public HOI detection benchmark V-COCO \cite{VSRL} dataset. Our code is available at \href{https://github.com/zhumanli/SGCN4HOI}{https://github.com/zhumanli/SGCN4HOI}.
\end{itemize}

\section{Related Works}
In HOI detection, deep learning methods have been widely used due to its rapid development and superiority in modelling complicated data. In this section, we review the highly relevant research which focuses on modelling the skeletal structure of humans and utilizing graph networks. %to boost HOI detection performance.

\subsection{Skeleton-based HOI Detection}
Human pose which provides fine-grained cues for HOI detection has been studied in existing works. Fang et al. \cite{PD-Net} propose a pairwise body-part attention module to guide the network focusing on important body parts of interaction detection. Wan et al. \cite{PMFNet} propose a zoom-in module to utilize the human skeleton features for mining interaction patterns between humans and objects. Generally, the human skeleton has been well studied in these methods, while the object skeleton and object keypoints representation are far less explored. Although Zheng et al. \cite{SIGN} apply the graph network to model the interactions between human joints and object keypoints, they simply use two corner points of the object bounding box. We argue that this is inadequate for object representation as object sizes and shapes greatly vary (e.g., a football is different from a horse in either size or structure). In this paper, we propose a novel skeleton-based object keypoints representation that benefits HOI detection by better capturing the underlying structure of the objects. This is motivated by the success of modelling humans using the skeletal structure in the literature.

\subsection{Graph Networks for Skeleton-based Action Recognition}
Graph neural netoworks have received increasing attention in skeleton-based action recognition due to the natural graphic structure of human body. It was firstly adopted in \cite{STGCN}, which represented human body joints as graph nodes to capture their correlations. Shi et al. \cite{AGCN} exploit the adaptive adjacency matrix with a fully connected graph to effectively learn the joint correlations. Zhang et al. \cite{SGN} propose to use the similarity between nodes as an adjacency matrix and inject semantics into human skeleton representation to guide network learning. In this paper, we bring in these insights of graph models in action recognition to HOI detection, that is we model both human keypoints and object keypoints as graph nodes to learn their fine-grained spatial correlations.

\begin{figure*}[htbp]
	\centering
	\includegraphics[width=\textwidth]{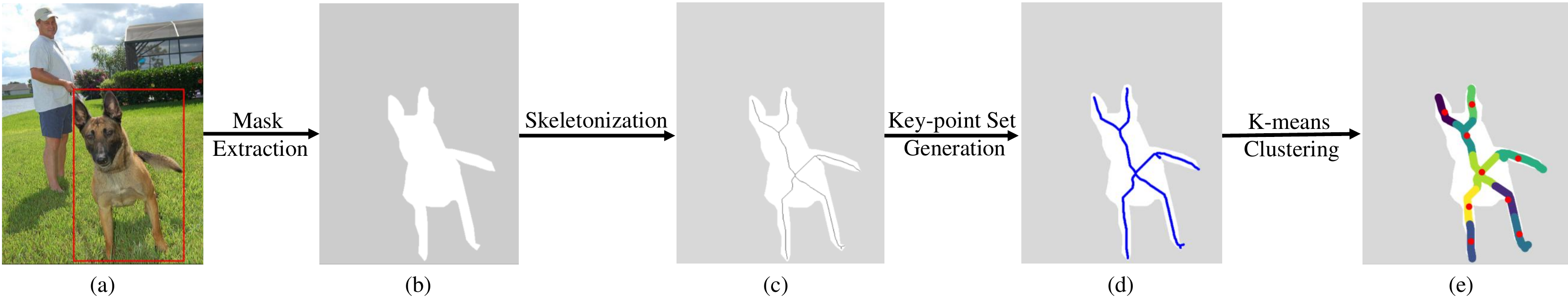}
	\caption{\footnotesize{Illustration of object keypoints extraction.}}
	\label{Fig: pointExtraction}
\end{figure*}

%\section{Overview}
%\section{Overall Framework}
\section{Overview of SGCN4HOI}
\label{Overview}
In this section, we give an overview of our proposed multi-stream framework for HOI detection. We start with %an overview of 
our model architecture (Section \ref{architecture}), followed by a brief introduction of the backbone VSGNet (\ref{vsgnet}). A detailed description of our proposed skeleton-aware graph convolution network will be presented in Section \ref{SpatialSkeletonStream}. 

%the spatial configuration stream (Section \ref{SpatialConfigurationStream}) and the visual stream (\ref{VisualStream}) that we adopted from VSGNet\cite{VSGNet}. Next, we give the detailed description of our proposed skeleton-aware graph convolution network (Section \ref{SpatialSkeletonStream}). 

\subsection{Model Architecture}
\label{architecture}
Our model for human-object interaction detection is a two-stage method that consists of two main steps: i) detecting instances and ii) predicting interactions for human-object pairs. First, human/object instances are detected by object detectors (e.g., Faster R-CNN \cite{Faster-R-CNN}). Second, we evaluate all the human-object pairs through the proposed spatial skeleton stream, the spatial configuration stream, and visual stream respectively, and the final HOI prediction is obtained by fusing the three stream's prediction scores. A graphical illustration of our framework is shown in Figure \ref{Fig: framework}.% show the overview of our framework.

The inputs of our model including bounding boxes $\mathbf{b}_h \in \mathbb{R}^4$ for human $h \in [1,H]$ and $\mathbf{b}_o \in \mathbb{R}^4$ for object $o \in [1,O]$, keypoints $\mathbf{v}_h \in \mathbb{R}^{17\times2}$ for human and $\mathbf{v}_o \in \mathbb{R}^{9\times2}$ for object. $H$ and $O$ denote the number of humans and objects respectively from a given image. Our main focus is to capture fine-grained spatial correlations between humans and objects by considering their keypoints as graph nodes. The bounding boxes are obtained from VSGNet \cite{VSGNet} which has included the instance detection using Faster R-CNN \cite{Faster-R-CNN}. Similar to previous works \cite{iCAN, TIN} that use ground-truth bounding boxes, we use the ground-truth keypoints of humans, and apply ground-truth segmentations of objects to extract their keypoints (details are decribed in Section \ref{PoinExtra}).

%present our proposed SGCN4HOI that utilizes keypoints from both human skeletons and object` skeletons for HOI detection. Below we start with an overview of our method (Section \ref{Overview}), followed by a brief introduction of the spatial configuration stream (Section \ref{SpatialConfigurationStream}) and the visual stream (\ref{VisualStream}) that we adopted from VSGNet\cite{VSGNet}. Next, we give the detailed description of our proposed skeleton-aware graph convolution network (Section \ref{SpatialSkeletonStream}). 

\subsection{Backbone VSGNet}
\label{vsgnet}
In order to take advantage of using available features to facilitate HOI detection, we adopt VSGNet \cite{VSGNet} as our backbone network to generate spatial configuration features and visual features through two streams:% . They are described as follows.

\subsubsection{Spatial Configuration Stream}
\label{SpatialConfigurationStream}
This stream aims to learn the coarse-grained spatial interactions between humans and objects from their spatial configuration patterns. The spatial configurations encode the spatial relationship of humans and objects that help visual features make better predictions \cite{VSGNet,iCAN}. Given the human bounding box $\mathbf{b}_h$ and object bounding box $\mathbf{b}_o$, two binary maps are firstly generated by allocating zeros to the outside area of the bounding box in the entire image. By stacking them, a 2-channel spatial configuration map $\mathbf{S}_{ho}$ is then generated for each human-object pair. 

The spatial configuration features $\mathbf{x}_{ho}^{Spa}$ of this stream are obtained by:
\begin{equation}
    \mathbf{x}_{ho}^{Spa}=\operatorname{VSGNet}\left(\mathbf{S}_{ho}\right)
\end{equation}

The interaction probabilities are then obtained:
\begin{equation}
    \mathbf{p}_{ho}^{Spa}=\sigma\left(\operatorname{FC}\left(\mathbf{x}_{ho}^{Spa}\right)\right)
\end{equation}
where $\sigma$ is the sigmoid function, $\operatorname{FC}$ is the fully connected network, and $\mathbf{p}_{ho}^{Spa}$ is the action class probabilities with the dimension that equals %size of 
the number of action categories.

\subsubsection{Visual Stream}
\label{VisualStream}
This stream extracts visual features for human-object pairs from images. It learns the human visual feature, object visual feature, and the global context (i.e., the entire image) feature from the human bounding box, object bounding box, and human-object union bounding box, respectively. Here, the context provides scene-specific global cues which are helpful for HOI prediction. 

The refined visual features by spatial configuration features are obtained:
\begin{equation}
    \mathbf{x}_{ho}^{Vis}=\operatorname{VSGNet}\left(\mathbf{b}_h,\mathbf{b}_o,\mathbf{x}_{ho}^{Spa}\right)
\end{equation}

We then obtain the interaction prediction $\mathbf{p}_{ho}^{Vis}$ of this stream as follows:
\begin{equation}
    \mathbf{p}_{ho}^{Vis}=\sigma\left(\operatorname{FC}\left(\mathbf{x}_{ho}^{Vis}\right)\right)
\end{equation}

%\subsection{Spatial Skeleton Stream}
\section{Skeleton-aware Graph Convolutional Network}
\label{SpatialSkeletonStream}
As the core of our framework, we propose to capture the fine-grained spatial correlations between humans and objects by using graph networks with keypoints of both of them. We first introduce the object keypoints representation, then the proposed skeleton-aware graph network will be presented. 

%\subsubsection{Object Keypoints Extraction}
\subsection{Object Keypoints Representation}
\label{PoinExtra}
We now introduce how do we extract object keypoints and it consists of three main steps. Given an image (Figure \ref{Fig: pointExtraction}(a)), we fist obtain the ground-truth mask of the object which is a dog in this example (Figure \ref{Fig: pointExtraction}(b)), we take the following steps to obtain its keypoints:
\begin{enumerate}[label=(\roman*)]
    \item \textbf{Skeleton Extraction.} We exploit a morphological skeletonization algorithm \cite{Zhang:skeleton84}  %library
    on segmentation to represent object skeletons, aiming to obtain the object structure information to facilitate the modelling of fine-grained spatial relations between the object and human. %which is not explored in existing works. %due to the limited literature on applying a unified algorithm for representing skeletons for various objects. 
    The idea is to utilize the well-studied object segmentation and available skeletonization algorithms to obtain an object's skeleton in an indirect manner. Concretely, we use the scikit-image library (in Python) that works on binary masks to generate object skeleton as shown in Figure \ref{Fig: pointExtraction}(c).
    
    \item \textbf{Key-point Set Generation.} The ending points and intersecting points of the skeleton are considered key points since they represent the topology of the shape of the object. %The ending points indicate the important edge information and the intersecting points denote the essential internal information, thus they can best represent an object's structure. 
    Same as \cite{keypoints}, an ending point is a pixel (point) with only one neighbor, and an intersecting point is a pixel with two or more neighbors. Specifically, we generate a $9\times9$ grid centered at each skeleton pixel and count the number of its adjacent pixels. By doing so, the key points are then extracted, which form a key-point set (dense blue points in Figure \ref{Fig: pointExtraction}(d)). 
    
    \item \textbf{Keypoints Extraction.} In order to better preserve the object skeleton structural information, we apply the K-means algorithm on the key-point set to obtain a certain number (9 in our experiment) of keypoints for an object. The K-means algorithm aims to partition the input data into k partitions (clusters) and can be used to get an intuition about the structure of the data. %Here we are trying to use it to obtain the object's skeleton structure, 
    Here we are trying to represent the object's skeletal structure using a spare set of keypoints. The K-means clustering is suitable for our goal of object structure representation. The reason is that it minimizes the intra-cluster variance of input skeletal points, the resulting finite number of cluster centers can well represent the local regions that are part of the object. As shown in Figure \ref{Fig: pointExtraction}(e)), by applying K-means, all the points are clustered into different groups (represented in different colors) with the cluster centers colored in red. %. The red points are the centers of each group, 
    %We 
    It can be seen that the distribution of the cluster centers can represent the underlying structure of the object, and they are stored as the final keypoints.
\end{enumerate}

%\subsubsection{Skeleton-aware Graph Convolution Network}
\subsection{Skeleton-aware Graph Convolution Network}
\label{skeleton}
We construct a spatial graph $G = (V, E)$ with the keypoints of humans and objects as graph nodes to learn their fine-grained structural connections. Similarities/affinities between nodes plus learnable weights are the edge connectivities during graph learning, such that we can have a fully connected graph to fully exploit the correlations between all the nodes, which has proved to be effective in action recognition \cite{AGCN,SGN}. 

\begin{figure}[htbp]
	\setlength{\abovecaptionskip}{6pt} 
	\centering
	\includegraphics[scale=0.6]{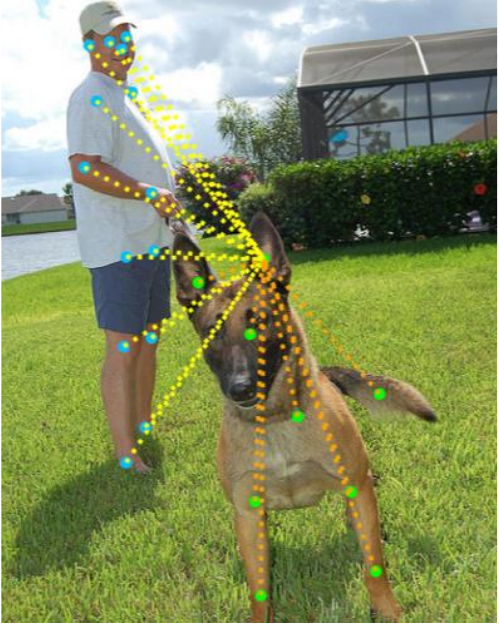}
	\caption{\footnotesize{Illustration of the graph structure. The connections of a single object keypoint is shown in this example for the sake of simplicity and readability.}}
	\label{graph}
\end{figure}

%Specifically, For human keypoints, 
Specifically, $V_{h}=\left\{ v_i|i=1, 2, 3,\cdots,J \right\}$ and  $V_{o}=\left\{ v_i|i=J+1, J+2, J+3,\cdots,J+K \right\}$ are the keypoint sets for human and object separately, where $J$ and $K$ are the number of keypoints of human and object respectively. The graph node set $V = \left\{V_h \cup V_o \right\}$, the edge set $E=\left\{ \varnothing \right\}$ is the initial input. For the sake of simplicity and readability, a simplified example is shown in Figure \ref{graph} by focusing on a single object keypoint. The object keypoint not only connects to its intra-class points (green points of this object, orange dotted connections) but also connects to all the inter-class points (light blue points of the human, yellow dotted connections). 
%As shown in Figure \ref{graph}, take a keypoint of the object as an illustration, it not only connects to its intra-class points (green points of this object, orange dotted connections) but also connects to all the inter-class points (red points of the human, blue dotted connections).

Given the keypoint locations of a human and an object, 
%$\mathbf{v}_h \in \mathbb{R}^{17\times2}$ and $\mathbf{v}_o \in \mathbb{R}^{9\times2}$ for a human and an object, 
we obtain their embeddings which are the inputs of graph convolutions using FC layers with ReLU activation function:
\begin{equation}
    \mathbf{f}_{in}=\operatorname{ReLU}\left(\operatorname{FC_2}\left(\operatorname{ReLU}\left(\operatorname{FC_1}\left(\mathbf{v}_{h/o}\right)\right)\right)\right)
\end{equation}

\begin{figure}[htbp]
	\setlength{\abovecaptionskip}{5pt} 
	\centering
	\includegraphics[scale=0.5]{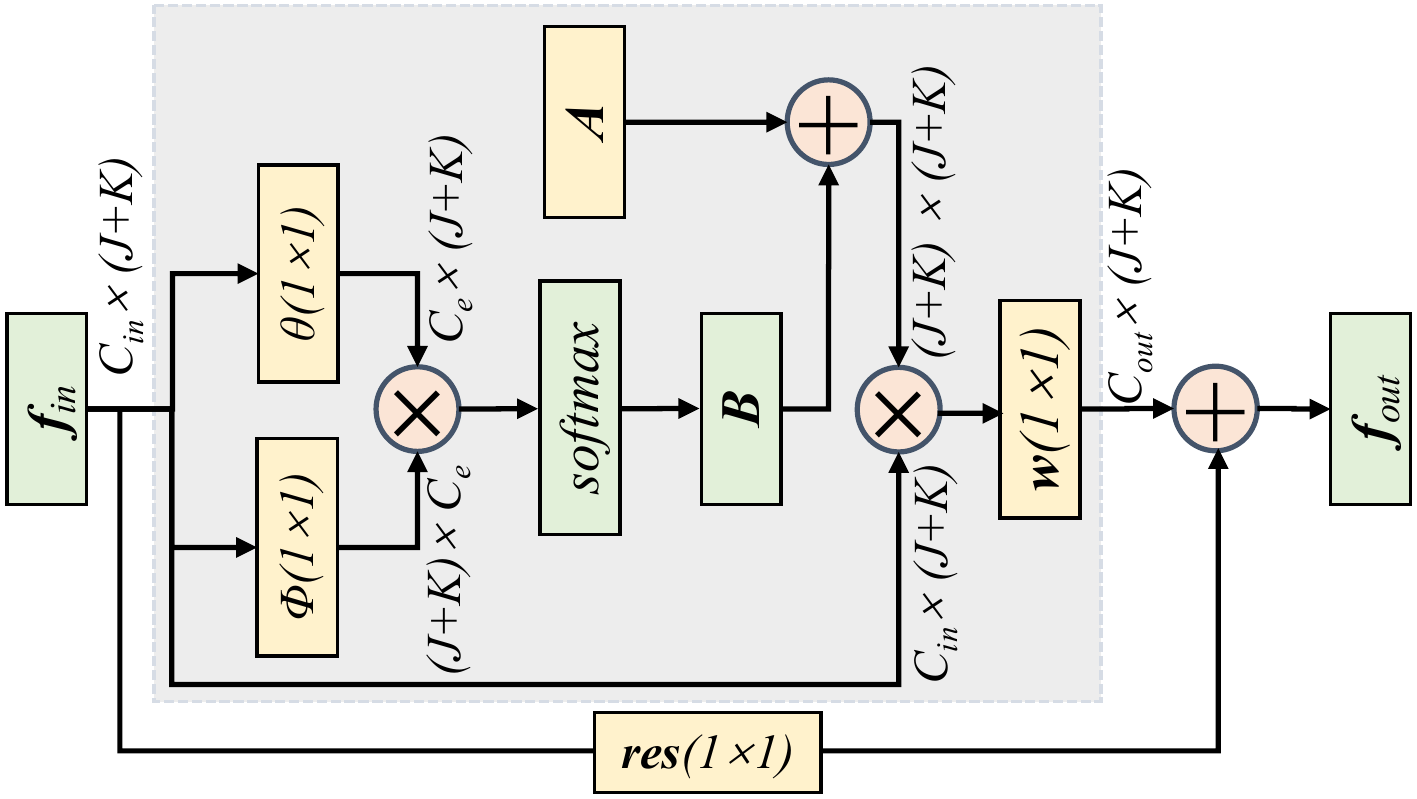}
	\caption{\footnotesize{Illustration of the graph convolution layer with residual connection.}}
	\label{Fig: gcnlayer}
\end{figure}

Once the embeddings of keypoints are obtained, we implement a residual graph convolution layer as follows, and its structure is shown in Figure \ref{Fig: gcnlayer}.
\begin{equation}
    \mathbf{f}_{out}=\operatorname{FC}\left(\mathbf{f}_{in}\right) + \operatorname{GCN}\left(\mathbf{f}_{in},\left( \mathbf{A} + \mathbf{B} \right)\right)
\end{equation}
where $\operatorname{FC}\left(\mathbf{f}_{in}\right)$ is the residual connection, $\operatorname{GCN}\left(\right)$ is the graph convolution operation. $(\mathbf{A+B})\in \mathbb{R}^{(J+K)\times (J+K)}$ is the adjacency matrix, in which $\mathbf{A}$ is fully learnable and its values are initialized as ones (i.e., adaptively adjust the values during learning), it can play the same role of the attention mechanism as it indicates the strength of the connections between two nodes. $\mathbf{B}$ is a data-dependent adjacency matrix represented by nodes similarities/affinities to determine whether there is a connection between two nodes and how strong the connection is \cite{AGCN}. The addition operation between $\mathbf{A}$ and $\mathbf{B}$ provides more flexibility than the commonly used dot multiplication on the adjacency matrix, that is if some elements of $\mathbf{A}$ are zeros, it still can generate connections by $\mathbf{B}$. The nodes similarities are defined as follows:
\begin{equation}
    s(i, j)=\theta\left(\mathbf{v}_{i}\right)^{T} \phi\left(\mathbf{v}_{j}\right)
\end{equation}
where $\theta$ and $\phi$ are transformation functions, which are implemented by an FC layer. We then stack three GCN layers to obtain the graph feature $\mathbf{x}_{ho}^{Graph}$. 

Visual image features can be helpful for identifying same/different objects, for instance, different cellphones often share similar colors and shapes, but they are different from bikes. %Similar to VSGNet \cite{VSGNet} that uses spatial configuration features to refine visual features, 
Thereby we use the visual features to refine our graph feature:
\begin{equation}
    \mathrm{x}_{ho}^{RG}=\mathbf{x}_{ho}^{Graph} \otimes \mathbf{x}_{ho}^{Vis}
\end{equation}

The interaction prediction $\mathbf{p}_{ho}^{Graph}$ of this stream is then made based on the refined graph feature:
\begin{equation}
    \mathbf{p}_{ho}^{Graph}=\sigma\left(\operatorname{FC}\left(\mathbf{x}_{ho}^{RG}\right)\right)
\end{equation}

Finally, we combine the predictions from each stream by multiplying the probabilities to make the final HOI prediction $\mathbf{p}_{ho}$ similar to previous works \cite{iCAN,InteractNet,TIN,VSGNet}:
\begin{equation}
    \mathbf{p}_{ho}=\mathbf{p}_{ho}^{Graph} \times \mathbf{p}_{ho}^{Vis} \times \mathbf{p}_{ho}^{Spa}
\end{equation}

\begin{figure*}[htbp]
	\centering
	\subfigure[Keypoints distribution represents object's underlying structure]{
		\begin{minipage}{0.47\linewidth}
			%\centering
			\includegraphics[width=\linewidth]{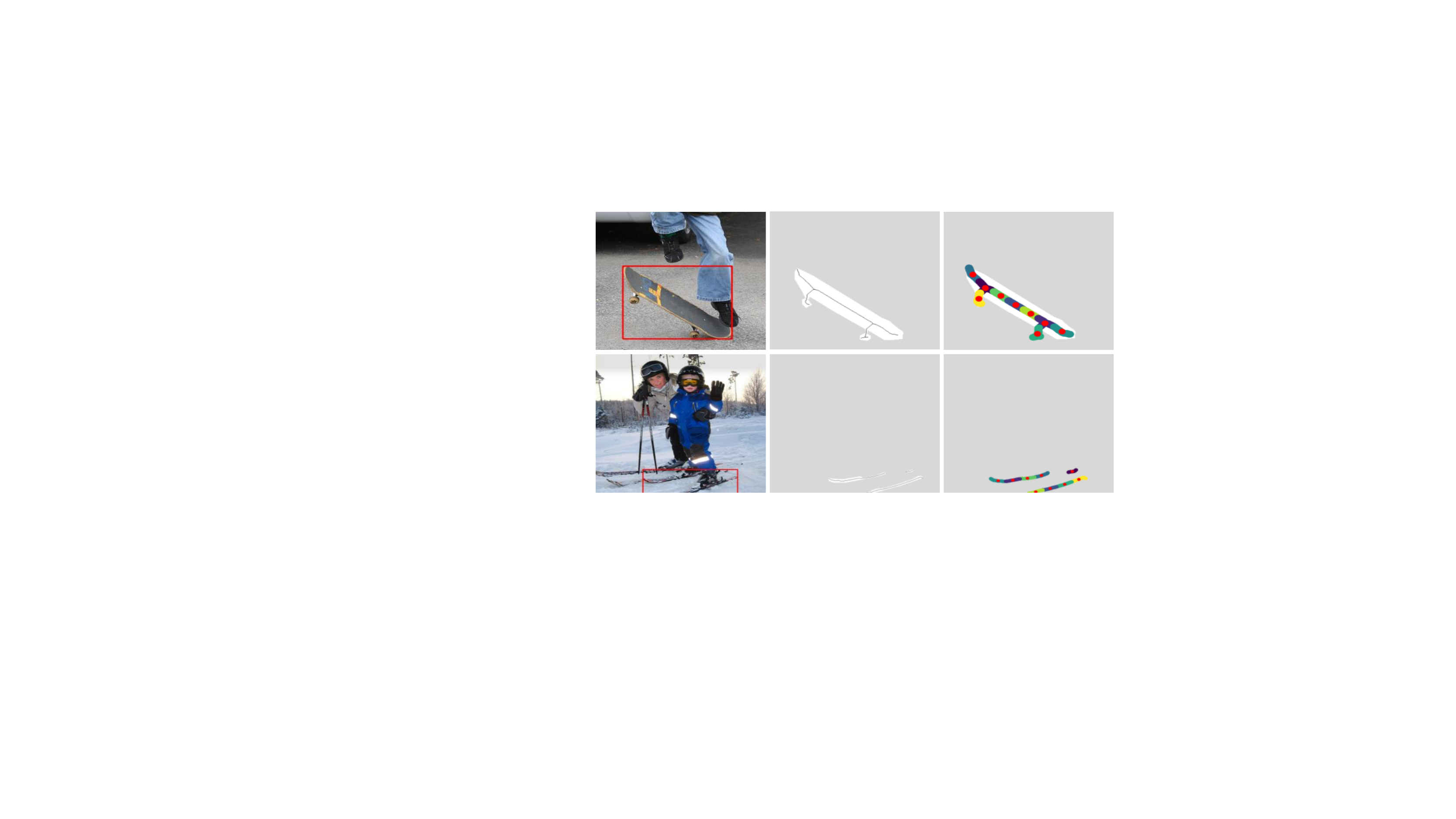}
		\end{minipage}
	}
    \quad
	\subfigure[Keypoints distribution fails for structure representation due to occlusion]{
		\begin{minipage}{0.47\linewidth}
			%\centering
			\includegraphics[width=\linewidth]{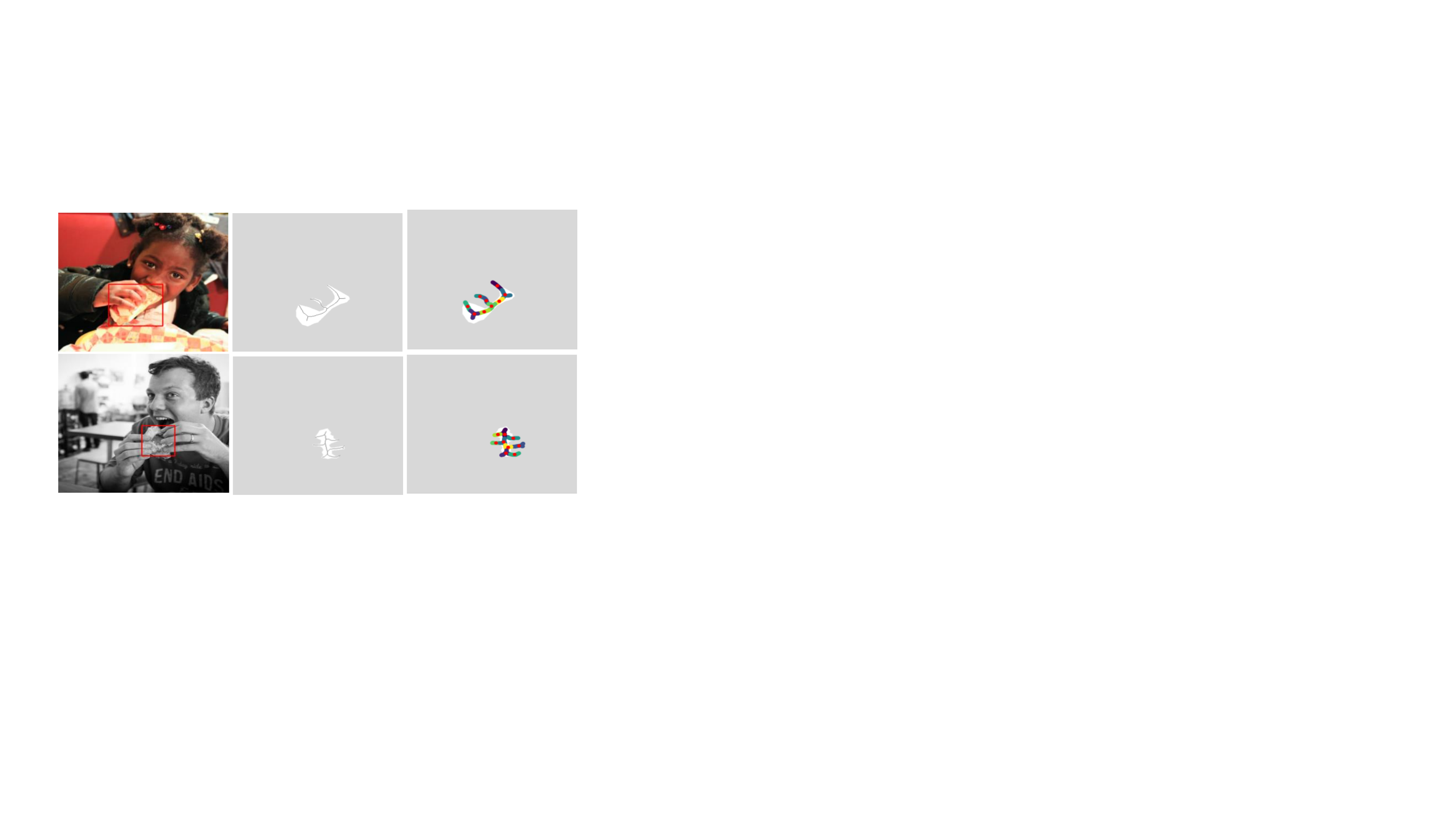} 
		\end{minipage}
	}
	\caption{Cases of keypoints distribution. Figures from left to right are: input image, object skeleton, keypoints with K-means clustering.} 
	\label{case} 
\end{figure*}

\section{Experiments}
\subsection{Dataset}
To evaluate the performance of our model, we conduct experiments in the HOI benchmark V-COCO \cite{VSRL} dataset, which is a subset of the COCO dataset \cite{COCO}. It has a total of 10,346 images with 16,199 human instances, 2533 images are for training, 2867 images are for validating, and 4946 images are for testing. Each human is annotated with binary labels indicating one of 29 different action categories. As four actions have no object associated with humans, and one action has limited samples, the results on the rest 24 classes are reported same as previous works \cite{VSRL,PMFNet,VSGNet}.

\subsection{Metrics}
We evaluate the performance on the commonly used role mean average precision (mAP) metrics: Scenario 1 and Scenario 2. The two scenarios indicate different evaluation methods for the cases when objects are occluded. In Scenario 1, the object bounding box must be empty, and in Scenario 2 the object bounding box is ignored. A prediction of a human-object pair is considered a true positive if both the predicted human and object bounding boxes have IoUs greater than 0.5 with ground-truth annotations and the interaction label is correct. 
%For the cases when objects are occluded, the object bounding box must be empty in Scenario 1, and in Scenario 2 the object bounding box is ignored. 

\subsection{Implementation Details}
Our output features of all three streams have the size of 512 dimensions, followed by an FC layer for the classification. The detected bounding boxes from Faster R-CNN \cite{Faster-R-CNN} are filtered by setting 0.6 confidence threshold for humans and 0.3 for objects.

We implement our network based on VSGNet \cite{VSGNet} and use Resnet-152 \cite{ResNet} as the backbone for visual feature extraction. We train our network on the V-COCO train and validation sets for 60 epochs with Stochastic Gradient Descent (SGD) optimizer, a batch size of 16, a learning rate of 0.005, a weight decay of 0.0001, and a momentum of 0.9. 

\subsection{Comparisons with the State of the Art}
We evaluate the performance of our proposed SGCN4HOI model and compare it with the state-of-the-art pose-based methods and other models in the HOI benchmark V-COCO dataset. We report the mean Average Precision (mAP) score in two settings provided in \cite{VSRL}.

Table \ref{AP} shows that our model %outperforms the existing pose-based models 
performs on par with or better than the pose-based methods in Scenario 1 and achieves state-of-the-art performance in Scenario 2. Compared to those methods, our model adds object structure information in addition to the human pose. By utilizing GCN to capture their fine-grained spatial interactions, the results show an improvement in the performance of HOI detection. %Compared to those methods, apart from the human pose, our model adds object structure information by utilizing GCN to capture their fine-grained spatial interactions, which turns out to be effective for HOI detection.

\begin{table}
	\centering
	\caption{Comparison of performance with pose-based methods. $AP_{role}^{\#1}$ and $AP_{role}^{\#2}$ represent the performance under Scenario1 and Scenario2 in V-COCO respectively.}
	\label{AP}
	%\resizebox{\linewidth}{!}{
	\begin{tabular}{cccc}\hline
		
		Pose-based Method & Backbone & $AP_{role}^{\#1}$ & $AP_{role}^{\#2}$ \\ \hline

		RPNN \cite{RPNN} & ResNet-50  & -\ & 47.5  \\
		TIN$^{*}$ \cite{TIN} & ResNet-50  & 48.7 & -\  \\
		PastaNet \cite{PastaNet} & ResNet-50 & 51.0 & 57.5  \\
		PMFNet \cite{PMFNet} & ResNet-50 & 52.0 & -\ \\
		PD-Net \cite{PD-Net} & ResNet-152  & 52.0 & -\ \\
		ACP$^{*}$ \cite{ACP} & ResNet-152 & 53.0 & -\ \\
		FCMNet \cite{FCMNet} & ResNet-50 & \textbf{53.1} & -\ \\ 
		SIGN \cite{SIGN} & ResNet50-FPN & \textbf{53.1} & -\ \\ \hline
		SGCN4HOI & ResNet-152  &  \textbf{53.1} & \textbf{57.9} \\ \hline
		
	\end{tabular}
	%}
\end{table}

Table \ref{AP-other} shows the comparison results with other methods. We observe that SGCN4HOI achieves an improvement of 1.3\% mAP and 0.9\% mAP on the backbone VSGNet \cite{VSGNet} in Scenario 1 and Scenario 2 respectively, and it outperforms all the other non-transformer methods as well. Compared to transformer-based methods \cite{HOITransformer,HOTR}, SGCN4HOI has much lower computational complexity, it also achieves better performance than HOI Transformer \cite{HOITransformer} in Scenario 1 and there was no Scenario 2 result reported in \cite{HOITransformer}.

\begin{table}
	\centering
	\caption{Comparison of performance with other methods. Our model achieves a good trade-off between performance and computational complexity comparing with transformer-based models.}
	\label{AP-other}
	\resizebox{\linewidth}{!}{
	\begin{tabular}{cccc}\hline
		
		Other Method & Backbone & $AP_{role}^{\#1}$ & $AP_{role}^{\#2}$ \\ \hline
		
		VSRL \cite{VSRL} & ResNet-50 & 31.8 & -\  \\
		InteractNet \cite{InteractNet} & ResNet-50  & 40.0 & 48.0  \\
		iCAN \cite{iCAN} & ResNet-50 & 45.3 & 52.4  \\
		VCL \cite{VCL} & ResNet-101 &  48.3 & -\  \\
		UnionDet \cite{UnionDet} & ResNet-50 & 47.5 & 56.2 \\
		IPNet \cite{IPNet} & Hourglass-104 & 51.0 & -\  \\ 
		DRG \cite{DRG} & ResNet-50 & 51.0 &  -\  \\
		VSGNet \cite{VSGNet} & ResNet-152 & 51.8 & 57.0  \\ \hline
		HOI Transformer \cite{HOITransformer} & ResNet-101  & 52.9 & -\  \\
		HOTR \cite{HOTR} (transformer method) & ResNet-50 & \textbf{55.2} & \textbf{64.4} \\
		\hline
		SGCN4HOI & ResNet-152  &  53.1 & 57.9 \\ \hline
		
	\end{tabular}
	}
\end{table}

In Table \ref{perclass}, we report the per-class performance of our model and compare it with methods that reported per-class AP. We can see that SGCN4HOI outperforms other methods in the majority of classes, particularly in ``skateboard-instr" and ``snowboard-instr'' classes. We believe that our proposed skeleton-based object keypoints representation well preserves such visible objects structures (shown in Figure \ref{case} (a)) which facilitates HOI detection. Notice that some of the classes such as ``eat-instr'' perform badly, and we believe this is mainly caused by two reasons. First, as claimed in VSGNet \cite{VSGNet}, the objects of class ``eat-instr'' are usually small and occluded in the images, thus object detectors often fail in these cases. Second, as shown in Figure \ref{case} (b), in these occluded cases, either the generated skeleton or keypoints cannot represent the object structure well due to the various occlusion configurations and the reliance on the quality of object segmentation.

\begin{table}[htbp]
	\centering
	\caption{Per class mAP comparisons with the existing methods in Scenario 1.}
	\label{perclass}
	\resizebox{\linewidth}{!}{
	\begin{tabular}{|c| c| c| c| c|}\hline
		HOI Class & InteractNet \cite{InteractNet} & iCAN \cite{iCAN} & VSGNet \cite{VSGNet} & SGCN4HOI \\ \hline
    	hold-obj &26.38 &29.06 & 48.27 & \textbf{51.55} \\ \hline
		sit-instr &19.88 & 26.04& \textbf{29.9} & 29.55 \\ \hline
		ride-instr &55.23 &61.9 & \textbf{70.84} & 68.9 \\ \hline
		look-obj & 20.2& 26.49& 42.78 & \textbf{47.56} \\ \hline
		hit-instr &62.32 &74.11 & 76.08 & \textbf{78.63} \\ \hline
		hit-obj &43.32 & 46.13& 48.6 & \textbf{50.62} \\ \hline
		eat-obj & 32.37& 37.73& 38.3 & \textbf{43.63} \\ \hline
    	eat-instr &1.97 & \textbf{8.26} & 6.3 & 3.18 \\ \hline
		jump-instr &45.14 & 51.45&  52.66 & \textbf{55.14} \\ \hline
		lay-instr &20.99 & \textbf{22.4} & 21.66 & 20.86 \\ \hline
		talk\_on\_phone & 31.77& 52.82& 62.23 & \textbf{63.55} \\ \hline
		carry-obj &33.11 &32.02 & \textbf{39.09} & 38.09 \\ \hline
		throw-obj & 40.44& 40.62& 45.12 & \textbf{53.09} \\ \hline
		catch-obj &42.52 & \textbf{47.61} & 44.84 & 46.87 \\ \hline
		cut-instr &22.97 &37.18 & 46.78 & \textbf{47.81} \\ \hline
		cut-obj &36.4 & 34.76& 36.58 & \textbf{37.64} \\ \hline
		work\_on\_comp &57.26 &56.29 & 64.6 & \textbf{69.11} \\ \hline
		ski-instr & 36.47& 41.46& \textbf{50.59} & 47.74 \\ \hline
		surf-instr &65.59 & 77.15& 82.22 & \textbf{82.81} \\ \hline
		skateboard-instr &75.51 & 79.35& 87.8 & \textbf{89.32} \\ \hline
		drink-instr &33.81 &32.19 & \textbf{54.41} & 47.99 \\ \hline
		kick-obj &69.44 & 66.89& 69.85 & \textbf{77.37} \\ \hline
		read-obj & 23.85&30.74 & \textbf{42.83} & 41.57 \\ \hline
		snowboard-instr & 63.85& 74.35& 79.9 & \textbf{81.35} \\ \hline
		Average & 40.0&45.3 & 51.8 & \textbf{53.1} \\ \hline
		
    	\end{tabular}}
\end{table}

\subsection{Ablation Studies}
% 1) network architecture (disable skeleton stream, disable human skeleton, disable object skeleton); 2) parameter analysis (k values)

\subsubsection{Analysis of the Skeleton Stream}
Our overall framework consists of three main streams. To evaluate how our proposed skeleton stream is affecting the overall performance, each component is evaluated. We consider the base model as the framework without skeleton stream, then the parts of human skeleton keypoints and object skeleton keypoints are evaluated separately. 

\begin{figure}[htbp]
	\centering
	\includegraphics[width=0.35\textwidth]{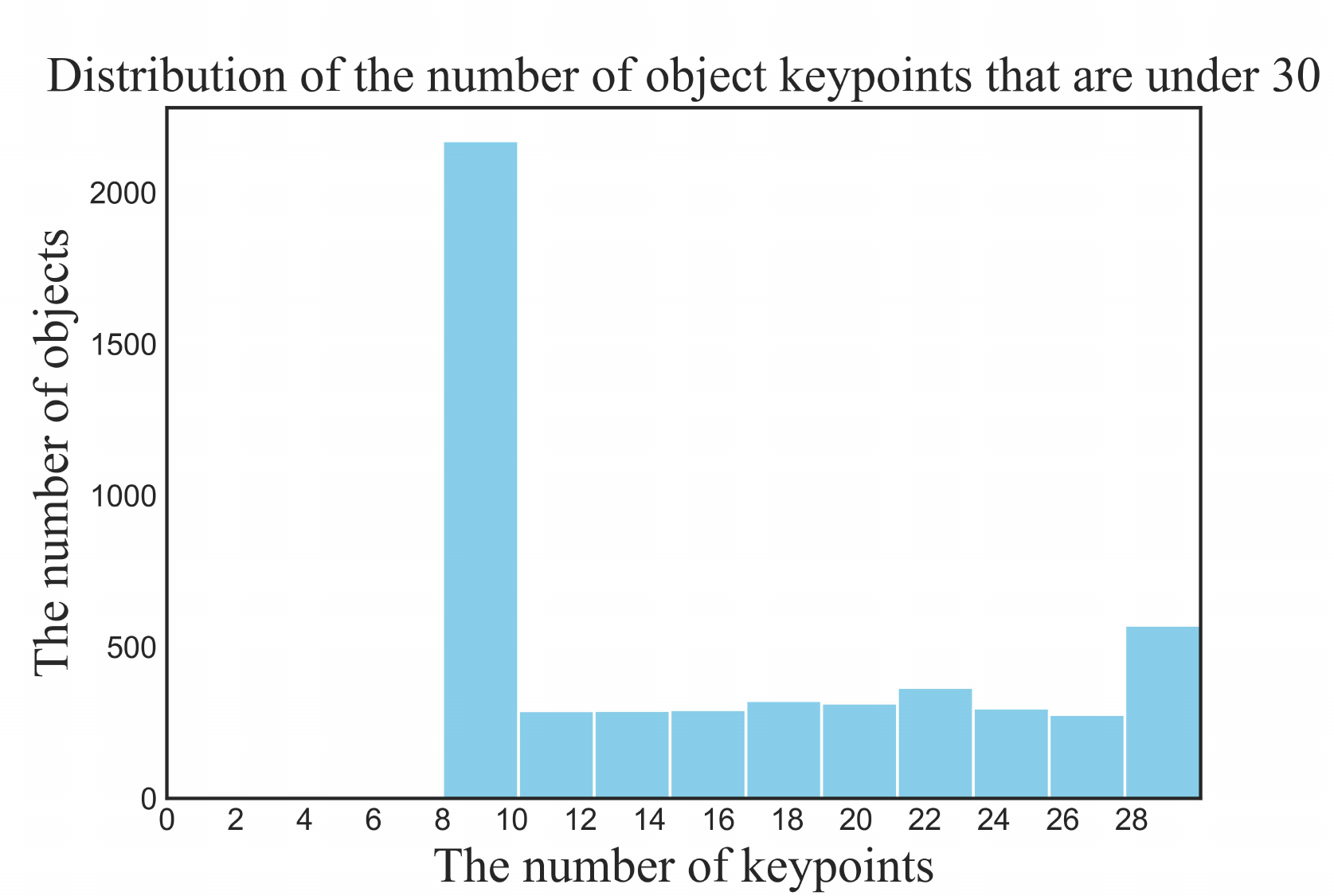}
	\caption{\footnotesize{Distribution of the number of object keypoints that are under 30.}}
	\label{Fig: statistics}
\end{figure}

The results are shown in Table \ref{component}. We can see that with the addition of object skeleton keypoints individually, the performance has decreased, while the performance of the addition of human skeleton keypoints alone increases, and adding both can further improve the performance. This demonstrates that the human skeleton is more important than the object skeleton for HOI detection, and the object skeleton can boost the performance by working together with the human skeleton. On the one hand, the human skeleton structure is unified and its poses are versatile, which provide more pose cues for action recognition, e.g., the skeleton of the ``sit" action is different from the ``lay" action. On the other hand, the object structures vary and the same object can be involved in multiple interactions, for example, a ``bed" can be involved in the ``sit" and ``lay" actions, in this case, it is hard for classifiers to differentiate such HOIs without human pose cues.

\begin{table}[htbp]
	\centering
	\caption{Components analysis of the skeleton stream.}
	\label{component}
	%\resizebox{\linewidth}{!}{
 	%\setlength{\tabcolsep}{10pt}
	\begin{tabular}{|c| c| c|}\hline
		Components & $AP_{role}^{\#1}$ & $AP_{role}^{\#2}$ \\ \hline
		Visual + Configuration (Base) & 50.5 & 55.1 \\ \hline
		Visual + Configuration + Skeleton (object) & 48.6 & 54.1 \\ \hline
		Visual + Configuration + Skeleton (human) & 52.3  & 57.2 \\ \hline
		Visual + Configuration + Skeleton (both) & \textbf{53.1} & \textbf{57.9} \\ \hline
    	\end{tabular}
    	%}
\end{table}

\subsubsection{Analysis of Parameter}
We analyze the affection of the number of object keypoints on the performance, that is the value k of K-means, and the results are shown in Table \ref{kvalue}. We can see that with the increase of the number of keypoints, the model performance improves gradually. We conclude that when the number of keypoints (e.g., 2 and 4) is too small, it is not enough to represent object structures, while larger numbers (e.g., 7 and 9) can represent object structures well.

In order to choose the best K parameter, we visualize the distribution of the number of object keypoints. % as shown in Figure \ref{Fig: statistics}. 
For the sake of clarity and readability, % we only show the distribution for the ranges of 0 to 20 (Figure \ref{Fig: statistics}(a)) and 0 to 40 (Figure \ref{Fig: statistics}(b)) as the largest number can be thousands. 
we only show the distribution for the range of 0 to 30 (Figure \ref{Fig: statistics}) as the largest number is a few thousands. It can be seen that most of the objects have a number of keypoints around 8 or 9, and we choose 9 in this work. We do not wish to have a very large number on K, because we padding zeros for small objects that have limited points to align with the K value, thereby a large K would harm the performance due to too many meaningless zeros.

\begin{table}[htbp]
	\centering
	\caption{K value analysis of object keypoints representation.}
	\label{kvalue}
	%\resizebox{\linewidth}{!}{
 	\setlength{\tabcolsep}{10pt}
	\begin{tabular}{|c| c| c|}\hline
		K value & $AP_{role}^{\#1}$ & $AP_{role}^{\#2}$ \\ \hline
		2 & 47.5 & 52.6 \\ \hline
		4 & 51.1 & 56.2 \\ \hline
		5 & 51.8 & 56.8 \\ \hline
		7 & 52.3 & 57.2\\ \hline
		9 & \textbf{53.1} & \textbf{57.9} \\ \hline
    	\end{tabular}
\end{table}

\section{Conclusion}
In this paper, we have proposed an effective skeleton-aware graph network SGCN4HOI that models fine-grained spatial correlations between human skeleton keypoints and object skeleton keypoints for HOI detection. Our proposed skeleton-based object keypoints representation method is able to preserve objects' sizes and structural information, which help differentiate objects that share subtle appearances or have different structures, resulting in better HOI detection performance. The presented multi-stream framework made full use of available features including visual features and spatial configuration features. Additionally, we found that visual features could be used as attention features to refine graph features that are learned from the skeleton stream. Experimental results have shown that SGCN4HOI improves the performance and outperforms the available pose-based state of the arts. 

In the future, more robust object keypoints representation can be explored, and the adaptive representation in object detection such as RepPoint \cite{RepPoints} and CFA \cite{CFA} can be considered. As discussed in the experimental results, our approach to object keypoints representation is limited by the quality of object segmentation, thereby it performs badly in occluded scenes. We hope our work can be helpful for future explorations on object keypoints representation.  

%\section*{Acknowledgement}

\bibliographystyle{IEEEtran}  
\bibliography{IEEEabrv,references}

\begin{thebibliography}{10}
\providecommand{\url}[1]{#1}
\csname url@samestyle\endcsname
\providecommand{\newblock}{\relax}
\providecommand{\bibinfo}[2]{#2}
\providecommand{\BIBentrySTDinterwordspacing}{\spaceskip=0pt\relax}
\providecommand{\BIBentryALTinterwordstretchfactor}{4}
\providecommand{\BIBentryALTinterwordspacing}{\spaceskip=\fontdimen2\font plus
\BIBentryALTinterwordstretchfactor\fontdimen3\font minus
  \fontdimen4\font\relax}
\providecommand{\BIBforeignlanguage}[2]{{%
\expandafter\ifx\csname l@#1\endcsname\relax
\typeout{** WARNING: IEEEtran.bst: No hyphenation pattern has been}%
\typeout{** loaded for the language `#1'. Using the pattern for}%
\typeout{** the default language instead.}%
\else
\language=\csname l@#1\endcsname
\fi
#2}}
\providecommand{\BIBdecl}{\relax}
\BIBdecl

\bibitem{VQA}
Y.~Goyal, T.~Khot, D.~Summers-Stay, D.~Batra, and D.~Parikh, ``Making the v in
  vqa matter: Elevating the role of image understanding in visual question
  answering,'' in \emph{Proceedings of the IEEE conference on computer vision
  and pattern recognition}, 2017, pp. 6904--6913.

\bibitem{ActivityNet}
F.~Caba~Heilbron, V.~Escorcia, B.~Ghanem, and J.~Carlos~Niebles, ``Activitynet:
  A large-scale video benchmark for human activity understanding,'' in
  \emph{Proceedings of the ieee conference on computer vision and pattern
  recognition}, 2015, pp. 961--970.

\bibitem{HOITransformer}
C.~Zou, B.~Wang, Y.~Hu, J.~Liu, Q.~Wu, Y.~Zhao, B.~Li, C.~Zhang, C.~Zhang,
  Y.~Wei, and J.~Sun, ``End-to-end human object interaction detection with hoi
  transformer,'' in \emph{2021 IEEE/CVF Conference on Computer Vision and
  Pattern Recognition (CVPR)}, 2021, pp. 11\,820--11\,829.

\bibitem{HOTR}
B.~Kim, J.~Lee, J.~Kang, E.-S. Kim, and H.~J. Kim, ``Hotr: End-to-end
  human-object interaction detection with transformers,'' in \emph{Proceedings
  of the IEEE/CVF Conference on Computer Vision and Pattern Recognition}, 2021,
  pp. 74--83.

\bibitem{QPIC}
M.~Tamura, H.~Ohashi, and T.~Yoshinaga, ``Qpic: Query-based pairwise
  human-object interaction detection with image-wide contextual information,''
  in \emph{2021 IEEE/CVF Conference on Computer Vision and Pattern Recognition
  (CVPR)}, 2021, pp. 10\,405--10\,414.

\bibitem{VSRL}
S.~Gupta and J.~Malik, ``Visual semantic role labeling,'' \emph{arXiv preprint
  arXiv:1505.04474}, 2015.

\bibitem{IPNet}
T.~Wang, T.~Yang, M.~Danelljan, F.~S. Khan, X.~Zhang, and J.~Sun, ``Learning
  human-object interaction detection using interaction points,'' in \emph{2020
  IEEE/CVF Conference on Computer Vision and Pattern Recognition (CVPR)}, 2020,
  pp. 4115--4124.

\bibitem{SG2HOI}
T.~He, L.~Gao, J.~Song, and Y.-F. Li, ``Exploiting scene graphs for
  human-object interaction detection,'' in \emph{Proceedings of the IEEE/CVF
  International Conference on Computer Vision}, 2021, pp. 15\,984--15\,993.

\bibitem{Faster-R-CNN}
S.~Ren, K.~He, R.~Girshick, and J.~Sun, ``Faster r-cnn: Towards real-time
  object detection with region proposal networks,'' in \emph{Proceedings of the
  28th International Conference on Neural Information Processing Systems -
  Volume 1}, ser. NIPS'15.\hskip 1em plus 0.5em minus 0.4em\relax Cambridge,
  MA, USA: MIT Press, 2015, p. 91–99.

\bibitem{detr}
N.~Carion, F.~Massa, G.~Synnaeve, N.~Usunier, A.~Kirillov, and S.~Zagoruyko,
  ``End-to-end object detection with transformers,'' in \emph{European
  conference on computer vision}.\hskip 1em plus 0.5em minus 0.4em\relax
  Springer, 2020, pp. 213--229.

\bibitem{VSGNet}
O.~Ulutan, A.~S.~M. Iftekhar, and B.~S. Manjunath, ``Vsgnet: Spatial attention
  network for detecting human object interactions using graph convolutions,''
  in \emph{2020 IEEE/CVF Conference on Computer Vision and Pattern Recognition
  (CVPR)}, 2020, pp. 13\,614--13\,623.

\bibitem{iCAN}
C.~Gao, Y.~Zou, and J.-B. Huang, ``ican: Instance-centric attention network for
  human-object interaction detection,'' \emph{arXiv preprint arXiv:1808.10437},
  2018.

\bibitem{Yao:CVPR2010}
B.~Yao and L.~Fei-Fei, ``Modeling mutual context of object and human pose in
  human-object interaction activities,'' in \emph{2010 IEEE Computer Society
  Conference on Computer Vision and Pattern Recognition}, 2010, pp. 17--24.

\bibitem{TIN}
Y.-L. Li, S.~Zhou, X.~Huang, L.~Xu, Z.~Ma, H.-S. Fang, Y.~Wang, and C.~Lu,
  ``Transferable interactiveness knowledge for human-object interaction
  detection,'' in \emph{2019 IEEE/CVF Conference on Computer Vision and Pattern
  Recognition (CVPR)}, 2019, pp. 3580--3589.

\bibitem{PMFNet}
B.~Wan, D.~Zhou, Y.~Liu, R.~Li, and X.~He, ``Pose-aware multi-level feature
  network for human object interaction detection,'' in \emph{2019 IEEE/CVF
  International Conference on Computer Vision (ICCV)}, 2019, pp. 9468--9477.

\bibitem{ETAI2022}
H.-B. Zhang, Y.-Z. Zhou, J.-X. Du, J.-L. Huang, Q.~Lei, and L.~Yang, ``Improved
  human-object interaction detection through skeleton-object relations,''
  \emph{Journal of Experimental \& Theoretical Artificial Intelligence},
  vol.~34, no.~1, pp. 41--52, 2022.

\bibitem{SIGN}
S.~Zheng, S.~Chen, and Q.~Jin, ``Skeleton-based interactive graph network for
  human object interaction detection,'' in \emph{2020 IEEE International
  Conference on Multimedia and Expo (ICME)}, 2020, pp. 1--6.

\bibitem{RepPoints}
Z.~Yang, S.~Liu, H.~Hu, L.~Wang, and S.~Lin, ``Reppoints: Point set
  representation for object detection,'' in \emph{2019 IEEE/CVF International
  Conference on Computer Vision (ICCV)}, 2019, pp. 9656--9665.

\bibitem{bbx-seg}
A.~Monroy and B.~Ommer, ``Beyond bounding-boxes: Learning object shape by
  model-driven grouping,'' in \emph{European Conference on Computer
  Vision}.\hskip 1em plus 0.5em minus 0.4em\relax Springer, 2012, pp. 580--593.

\bibitem{CFA}
Z.~Guo, C.~Liu, X.~Zhang, J.~Jiao, X.~Ji, and Q.~Ye, ``Beyond bounding-box:
  Convex-hull feature adaptation for oriented and densely packed object
  detection,'' in \emph{2021 IEEE/CVF Conference on Computer Vision and Pattern
  Recognition (CVPR)}, 2021, pp. 8788--8797.

\bibitem{ReDet}
J.~Han, J.~Ding, N.~Xue, and G.-S. Xia, ``Redet: A rotation-equivariant
  detector for aerial object detection,'' in \emph{2021 IEEE/CVF Conference on
  Computer Vision and Pattern Recognition (CVPR)}, 2021, pp. 2785--2794.

\bibitem{Zhang:skeleton84}
T.~Y. Zhang and C.~Y. Suen, ``A fast parallel algorithm for thinning digital
  patterns,'' \emph{Commun. ACM}, vol.~27, no.~3, p. 236–239, mar 1984.

\bibitem{PD-Net}
X.~Zhong, C.~Ding, X.~Qu, and D.~Tao, ``Polysemy deciphering network for
  human-object interaction detection,'' in \emph{European Conference on
  Computer Vision}.\hskip 1em plus 0.5em minus 0.4em\relax Springer, 2020, pp.
  69--85.

\bibitem{STGCN}
S.~Yan, Y.~Xiong, and D.~Lin, ``Spatial temporal graph convolutional networks
  for skeleton-based action recognition,'' in \emph{Thirty-second AAAI
  conference on artificial intelligence}, 2018.

\bibitem{AGCN}
L.~Shi, Y.~Zhang, J.~Cheng, and H.~Lu, ``Two-stream adaptive graph
  convolutional networks for skeleton-based action recognition,'' in \emph{2019
  IEEE/CVF Conference on Computer Vision and Pattern Recognition (CVPR)}, 2019,
  pp. 12\,018--12\,027.

\bibitem{SGN}
P.~Zhang, C.~Lan, W.~Zeng, J.~Xing, J.~Xue, and N.~Zheng, ``Semantics-guided
  neural networks for efficient skeleton-based human action recognition,'' in
  \emph{2020 IEEE/CVF Conference on Computer Vision and Pattern Recognition
  (CVPR)}, 2020, pp. 1109--1118.

\bibitem{keypoints}
G.~Rojas-Albarrac{\'\i}n, C.~A. Carbajal, A.~Fern{\'a}ndez-Caballero, and M.~T.
  L{\'o}pez, ``Skeleton simplification by key points identification,'' in
  \emph{Mexican Conference on Pattern Recognition}.\hskip 1em plus 0.5em minus
  0.4em\relax Springer, 2010, pp. 30--39.

\bibitem{InteractNet}
G.~Gkioxari, R.~Girshick, P.~Dollár, and K.~He, ``Detecting and recognizing
  human-object interactions,'' in \emph{2018 IEEE/CVF Conference on Computer
  Vision and Pattern Recognition}, 2018, pp. 8359--8367.

\bibitem{COCO}
T.-Y. Lin, M.~Maire, S.~Belongie, J.~Hays, P.~Perona, D.~Ramanan,
  P.~Doll{\'a}r, and C.~L. Zitnick, ``Microsoft coco: Common objects in
  context,'' in \emph{European conference on computer vision}.\hskip 1em plus
  0.5em minus 0.4em\relax Springer, 2014, pp. 740--755.

\bibitem{ResNet}
K.~He, X.~Zhang, S.~Ren, and J.~Sun, ``Deep residual learning for image
  recognition,'' in \emph{Proceedings of the IEEE conference on computer vision
  and pattern recognition}, 2016, pp. 770--778.

\bibitem{RPNN}
P.~Zhou and M.~Chi, ``Relation parsing neural network for human-object
  interaction detection,'' in \emph{2019 {IEEE/CVF} International Conference on
  Computer Vision, {ICCV} 2019, Seoul, Korea (South), October 27 - November 2,
  2019}.\hskip 1em plus 0.5em minus 0.4em\relax {IEEE}, 2019, pp. 843--851.

\bibitem{PastaNet}
Y.-L. Li, L.~Xu, X.~Liu, X.~Huang, Y.~Xu, S.~Wang, H.-S. Fang, Z.~Ma, M.~Chen,
  and C.~Lu, ``Pastanet: Toward human activity knowledge engine,'' in
  \emph{2020 IEEE/CVF Conference on Computer Vision and Pattern Recognition
  (CVPR)}, 2020, pp. 379--388.

\bibitem{ACP}
D.-J. Kim, X.~Sun, J.~Choi, S.~Lin, and I.~S. Kweon, ``Detecting human-object
  interactions with action co-occurrence priors,'' in \emph{European Conference
  on Computer Vision}.\hskip 1em plus 0.5em minus 0.4em\relax Springer, 2020,
  pp. 718--736.

\bibitem{FCMNet}
Y.~Liu, Q.~Chen, and A.~Zisserman, ``Amplifying key cues for
  human-object-interaction detection,'' in \emph{European Conference on
  Computer Vision}.\hskip 1em plus 0.5em minus 0.4em\relax Springer, 2020, pp.
  248--265.

\bibitem{VCL}
Z.~Hou, X.~Peng, Y.~Qiao, and D.~Tao, ``Visual compositional learning for
  human-object interaction detection,'' in \emph{European Conference on
  Computer Vision}.\hskip 1em plus 0.5em minus 0.4em\relax Springer, 2020, pp.
  584--600.

\bibitem{UnionDet}
B.~Kim, T.~Choi, J.~Kang, and H.~J. Kim, ``Uniondet: Union-level detector
  towards real-time human-object interaction detection,'' in \emph{European
  Conference on Computer Vision}.\hskip 1em plus 0.5em minus 0.4em\relax
  Springer, 2020, pp. 498--514.

\bibitem{DRG}
C.~Gao, J.~Xu, Y.~Zou, and J.-B. Huang, ``Drg: Dual relation graph for
  human-object interaction detection,'' in \emph{European Conference on
  Computer Vision}.\hskip 1em plus 0.5em minus 0.4em\relax Springer, 2020, pp.
  696--712.

\end{thebibliography}

\end{document}